# Hybrid SIFT-SNN for Efficient Anomaly Detection of Traffic Flow-Control Infrastructure

Munish Rathee
*School of Engineering, Computer and Mathematical Science (of Auckland University of Technology)*
Auckland, New Zealand
munish.rathee@autuni.ac.nz

Boris Bačić
*School of Engineering, Computer and Mathematical Science (of Auckland University of Technology)*
*Institute of Biomedical Technologies (IBTec)*
Auckland, New Zealand
boris.bacic@aut.ac.nz

Maryam Doborjeh
*Knowledge Engineering and Discovery Research Institute (KEDRI) (of Auckland University of Technology)*
Auckland, New Zealand
maryam.doborjeh@aut.ac.nz

*Abstract*— This paper presents the SIFT-SNN framework, a low-latency neuromorphic signal-processing pipeline for real-time detection of structural anomalies in transport infrastructure. The proposed approach integrates Scale-Invariant Feature Transform (SIFT) for spatial feature encoding with a latency-driven spike conversion layer and a Leaky Integrate-and-Fire (LIF) Spiking Neural Network (SNN) for classification. The Auckland Harbour Bridge dataset is recorded under various weather and lighting conditions, comprising 6,000 labelled frames that include both real and synthetically augmented unsafe cases. The presented system achieves a classification accuracy of 92.3% ± 0.8% with a per-frame inference time of 9.5 ms. Achieved sub-10 millisecond latency, combined with sparse spike activity (8.1%), enables real-time, low-power edge deployment. Unlike conventional CNN-based approaches, the hybrid SIFT-SNN pipeline explicitly preserves spatial feature grounding, enhances interpretability, supports transparent decision-making, and operates efficiently on embedded hardware. Although synthetic augmentation improved robustness, generalisation to unseen field conditions remains to be validated. The SIFT-SNN framework is validated through a working prototype deployed on a consumer-grade system and framed as a generalisable case study in structural safety monitoring for movable concrete barriers, which, as a traffic flow-control infrastructure, is deployed in over 20 cities worldwide.

*Keywords—Spiking Neural Network, neuromorphic computing, signal processing, edge AI, anomaly detection, infrastructure monitoring, low-latency inference*

## I. Introduction

Structural defects in transport infrastructure contribute to over 34% of road accidents globally [1]. Movable Concrete Barrier (MCB) systems, deployed in more than 20 cities worldwide, enable dynamic lane allocation on critical bridges and highways. The Auckland Harbour Bridge (AHB) carries more than 154,000 vehicles daily, with peak traffic exceeding 200,000 [2,3]. To manage daily traffic flow control, the MCB segments, each weighing 750 kg, are repositioned using NZD 1.4 million Barrier Transfer Machines (BTMs) [2, 3]. In addition to the operational costs, the integrity of the entire movable lane relies on metal safety pins that interconnect the MCB segments. Since the system's installation on AHB in 1990, MCB inspections have relied on manual protocols that require personnel to traverse over a kilometre under live traffic and varied environmental conditions, posing safety risks and operational inefficiencies (Fig. 1).

The early proof-of-concept work using deep convolutional neural networks (CNNs) showed promise in automating MCB pin inspection [4], but it also required computationally intensive architectures, restricting deployment in embedded or energy-constrained environments. Building on this foundation, this study proposes a hybrid neuromorphic vision framework integrating Scale-Invariant Feature Transform (SIFT) descriptors [5] for spatial feature extraction with a latency-driven spike encoding layer and a Leaky Integrate-and-Fire (LIF) Spiking Neural Network (SNN) for classification. This combination enables biologically inspired computation with improved interpretability, as spike activations correspond directly to spatially grounded SIFT features.

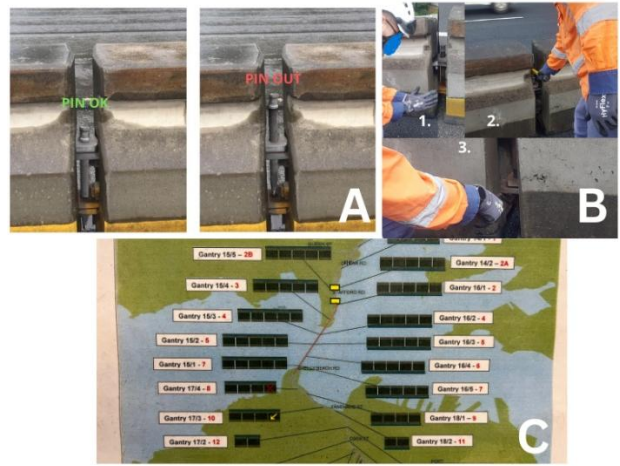

Fig. 1. Manual inspection process for metal pins securing the Auckland Harbour Bridge's movable concrete barrier (MCB) system: (A) identification of pins in either secure (OK) or dislodged (Out) states; (B) on-site detection and manual reinsertion of displaced pins while traffic remains active; (C) geographic distribution of MCB segments across the bridge [4].

While CNNs excel at hierarchical feature representation, their high energy demands and limited transparency hinder practical deployment in embedded infrastructure monitoring systems [6, 7]. In contrast, SNNs perform sparse, event-driven computation, enabling low-latency inference and reduced power consumption, while maintaining temporal fidelity [8]. Most SNN approaches rely on raw event-camera input or coarse frame-wide encodings. The proposed method bridges this gap by converting semantically meaningful SIFT descriptors into spike trains that preserve spatial structure, enabling efficient and interpretable classification.

The key contributions are as follows:

1. A novel SIFT-SNN hybrid model that combines spatial interpretability with temporal efficiency for anomaly detection in transport infrastructure.
2. A latency-based spike encoding mechanism that preserves descriptor-level spatial information.
3. Achieved SIFT-SNN accuracy (92.3% ± 0.8%) and an inference time of ~9.5 ms/frame, with improved deployability compared to CNN baselines.

The remainder is organised as follows: Section 2 reviews related work; Section 3 describes the methodology; Section 4 presents experiments and results; and Section 5 concludes with a discussion and future directions.

## II. RELATED WORK

Automated inspection of transport infrastructure has increasingly leveraged hybrid machine learning and computer vision systems to reduce manual labour [4, 9], enhance operational safety [10, 11], and enable real-time decision-making [10]. While these methods offer advanced inspection capabilities, reinforcement learning and large deep learning models remain computationally intensive [10, 12], limiting their feasibility for real-time and large-scale deployment.

Convolutional neural networks (CNNs) have been widely adopted for anomaly detection, structural damage classification, and object recognition in infrastructure monitoring [11, 13, 14]. Although CNNs achieve strong performance by learning complex hierarchical features, deployment in operational environments is constrained by high computational demand, limited interpretability, and susceptibility to domain-specific variations [15].

Before the adoption of deep learning, traditional methods such as Scale-Invariant Feature Transform (SIFT), Histogram of Oriented Gradients (HOG), and Canny edge detection provided geometric robustness and computational efficiency [15-17]. However, these hand-crafted approaches struggled with generalisation in unstructured, dynamic outdoor scenes [18]. This limitation has driven interest in neuromorphic computing, where Spiking Neural Networks (SNNs) perform sparse, event-driven computation inspired by biological neurons [19-21]. The presented asynchronous processing model enables low-latency inference and reduced energy consumption, making SNNs highly suited to embedded and edge-based infrastructure monitoring [22].

Despite their potential, most SNN applications in vision tasks rely either on raw event-based camera input or on converting entire image frames into spike trains using generic, frame-wide encoding strategies [8]. The integration of semantically rich descriptors, such as SIFT keypoints, into temporally structured spike representations remains unexplored, mainly in infrastructure safety applications.

This gap is addressed by introducing a hybrid SIFT-SNN architecture that:

1. Encodes interpretable, local visual descriptors into latency-coded spike trains;
2. Preserves spatial structure in the spike domain; and
3. Classifies anomalies using a low-latency Leaky Integrate-and-Fire (LIF) network.

The SIFT-SNN combination aims to strike a balance between semantic precision, computational efficiency, and deployment feasibility, enabling real-time safety monitoring for mobile infrastructure systems, such as movable barrier inspection.

## III. METHODOLOGY

### A. Data Acquisition and Pre-processing

High-resolution video data collected from a movable concrete barrier (MCB) inspection system deployed on the Auckland Harbour Bridge (AHB) as a representative case study for transport infrastructure safety monitoring. Data capture is performed using a multisensor experimental rig mounted on the Barrier Transfer Machine (BTM), equipped with GoPro cameras, an iPhone 13 Pro, a Samsung A7 mobile device, and an Apple iPad 6, powered by an external battery pack. The configuration enabled multi-angle, high-frame-rate acquisition under varied operational conditions, including low-light, drizzle, and overcast conditions, with a focus on the metal pins that secure 750 kg MCB segments. Figure 2 illustrates the BTM-mounted hardware configuration, showing camera placements and auxiliary sensors. The system supported synchronised capture and geospatial annotation of MCB segments in live traffic conditions.

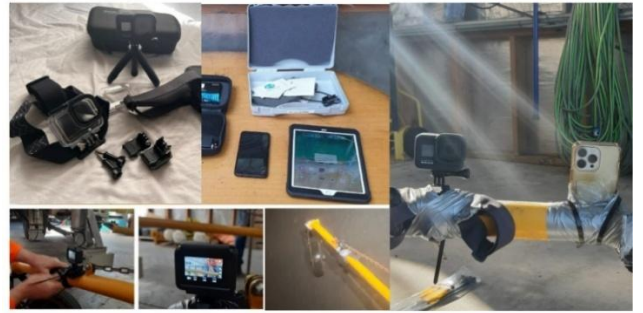

Fig. 2. Data collection setup mounted on the Barrier Transfer Machine (BTM), showing: iPhone 13 Pro, Samsung A7 mobile, Apple iPad 6, external power bank, and GoPro cameras. The shown equipment configuration enabled high-frame-rate recording and geospatial annotation of MCB segments under varied environmental conditions.

Video footage recorded at 120 fps and uniformly downsampled to 30 fps to balance temporal resolution with computational efficiency. For each frame, a fixed Region of Interest (ROI) is extracted around each pin location. Frames are then converted to grayscale, histogram-equalised to enhance contrast, and normalised to zero mean and unit variance. The pre-processing yielded 4,500 labelled Pin_OK frames representing correctly seated pins.

Due to operational policy and safety restrictions, AHB staff did not permit the recording of deliberately displaced pins. Although unsafe pin positions are occasionally observed during routine commutes, controlled filming is not authorised. As a result, generating synthetic unsafe data became the only feasible option.

The synthetic data generation approach also has broader implications:

It provides a methodology for developing models in safety-critical infrastructure contexts where the minority class is extremely scarce or capturing it is prohibited by management or security policies.

To construct the unsafe class (Pin_OUT), 120 base frames are manually created by digitally removing pins and overlaying dislodged or occluded templates in Adobe Photoshop. These expanded to 1,500 frames through augmentation techniques, including perspective warping, illumination variation, rotation (±10°), occlusion overlays, positional jitter, and morphological distortion. The synthetic generation pipeline (Fig. 3) simulates realistic visual degradations that may occur during operation, while (Fig. 4) illustrates representative augmented samples.

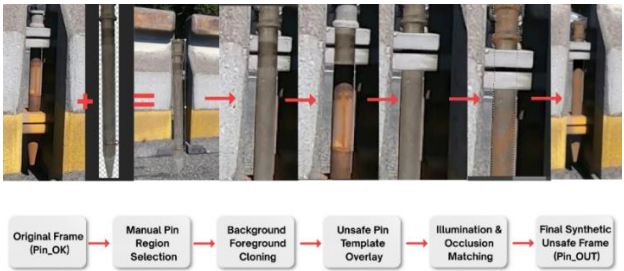

Fig. 3. Synthetic unsafe frame generation pipeline. Safe ROIs are edited and augmented to simulate unsafe pin scenarios under realistic visual conditions [9].

Figure 4 presents examples of augmentation techniques applied to the unsafe class, showcasing the diversity introduced to support generalisation.

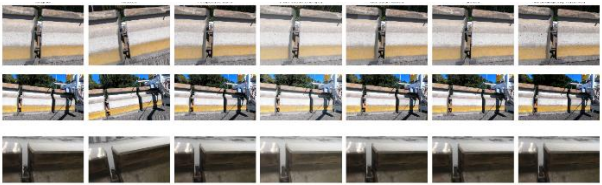

Fig. 4. Example augmentations: original frame, rotation, perspective warp, gamma shift, occlusion, translation, and shape distortion. Augmentation improves diversity under deployment-like variations.

The final dataset comprised 6,000 annotated frames with a 3:1 class ratio (Table 1). The dataset served as the input to the feature extraction stage, where each ROI is processed using the Scale-Invariant Feature Transform (SIFT) algorithm to produce descriptor vectors for spike encoding.

TABLE 1: DATASET COMPOSITION FOLLOWING PREPROCESSING AND AUGMENTATION

| Class | Source | Frames | Notes |
|---|---|---|---|
| Pin_OK | Real-world video | 4,500 | Sampled from a 30fps downsampled video. |
| Pin_OUT | Synthetic + Augmented | 1,500 | Based on 120 base frames + augmentation. |

## B. SIFT Keypoint Extraction and Descriptor Encoding

Each pre-processed ROI is transformed using the Scale-Invariant Feature Transform (SIFT) [5], implemented in Python using OpenCV [10]. SIFT is selected for its robustness to scale, rotation, and illumination variation, as well as its ability to localise features in structured visual anomaly detection tasks. This makes it particularly suitable for infrastructure safety inspection, where lighting, perspective, and environmental conditions vary during operation.

The extraction pipeline began with grayscale normalisation and histogram equalisation to enhance contrast, followed by keypoint detection using cv2.SIFT_create(). IFT identifies extrema in scale space, computes orientation histograms, and generates 128-dimensional descriptors for each keypoint. To maintain a consistent input size for downstream processing, keypoints ranked by response magnitude, and the top $N = 100$ are retained per frame. This yielded a fixed descriptor vector of $100 \times 128 = 12,800$ elements. Frames with fewer than 100 keypoints are zero-padded to ensure dimensional consistency without introducing artificial features.

Preservation of spatial structure is achieved by retaining the original keypoint ordering based on their pixel coordinates before vector concatenation. This ordering is essential for mapping descriptors into spike trains in the subsequent encoding stage, ensuring that spatial neighbourhoods in the image domain are reflected in temporal proximity in the spike domain. Class-wise inspection revealed consistent keypoint clusters around structural elements, such as pin edges, anchor points, and surrounding concrete textures—regions where anomalies are most likely to occur.

Figure 5 presents a side-by-side comparison of SIFT keypoints for Pin_OK and Pin_OUT samples, illustrating both the full-keypoint distribution and the top 100 subset. The term structural divergence refers to the distinct spatial clustering and density of SIFT keypoints between secure and dislodged pin states, reflecting geometric differences in local texture and contour features that support class separation.

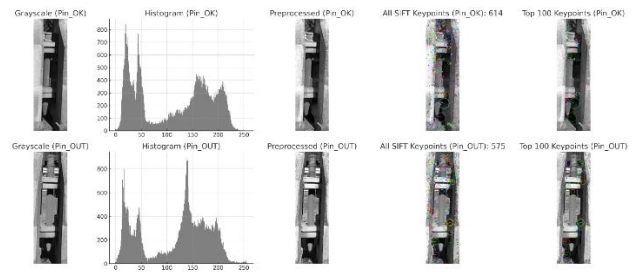

Fig. 5. SIFT processing comparison for Pin_OK vs. Pin_OUT: (1) grayscale ROI, (2) intensity histogram, (3) equalised image, (4) all keypoints, and (5) top 100 keypoints. Clear structural divergence supports class separation.

While initial attempts are made to implement the process in MATLAB using VLFeat [25], support limitations and compatibility issues led to the adoption of a fully Python-based workflow for stability, reproducibility, and future integration with spiking neural network toolchains.

## C. SIFT Descriptor as Spike Encoding Strategy

The 12,800-dimensional SIFT descriptor vector extracted from each ROI is first L2-normalised to the [0, 1] range and then converted into spike timings using a latency (time-to-first-spike) coding scheme. A fixed simulation window of $T = 100ms$ allocated per frame, reflecting the temporal resolution at which the downstream SNN processed input events. In this scheme, features with higher magnitudes are assigned earlier spike times, modelling the principle that stronger sensory stimuli elicit faster neural responses in biological systems.

The latency $t_i$ for the $i^{th}$ descriptor value $x_i$ computed as:

$$t_i = T \cdot (1 - x_i) \qquad (1)$$

Where $T$ is the total simulation window and $x_i$ is the normalised descriptor value. This formulation ensured that each input channel generated a single spike, producing a sparse, temporally ordered representation.

Spatial information preservation is achieved by retaining the original ordering of SIFT descriptors based on keypoint coordinates before vector concatenation. Consequently, descriptor neighbourhoods in the spatial domain are mapped to temporally adjacent spikes in the encoded stream. The spatial–temporal correspondence is crucial for enabling the SNN to exploit both feature salience and relative positioning.

The latency encoding yielded an event stream that is both compact and energy-efficient, as it emits only one spike per input dimension. This reduced the average spike activity to 8.1% across the simulation window, directly supporting low-power operation on edge hardware. Moreover, the fixed temporal window enabled deterministic inference latency, a crucial consideration for real-time infrastructure monitoring systems.

Figure 6 presents latency histograms, spike matrices, and spike latency heatmaps for Pin_OK and Pin_OUT frames, visualising the structural divergence encoded in temporal dynamics.

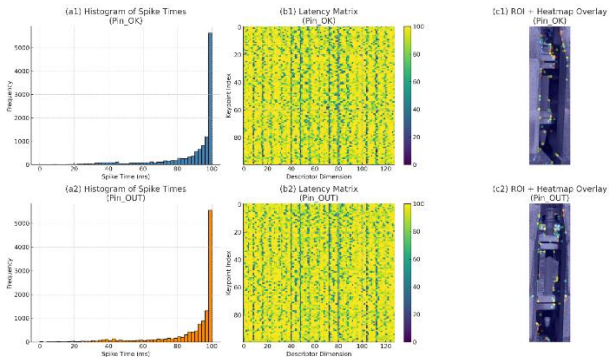

Fig. 6. Latency encoding visualisation for Pin_OK and Pin_OUT frames. (a1, a2) Histograms of latency-encoded spike times derived from the top 100 SIFT descriptors. (b1, b2) Latency-encoded spike time matrices (100 × 128), with colour-coded spike timing in milliseconds. (c1, c2) ROI overlays with spike latency heat maps; red/yellow denote early spikes, and blue regions indicate delayed activation. Distinct temporal salience patterns emerge across classes, reflecting structural variations in the pin region.

The distinct temporal salience patterns emerging from this encoding highlight its ability to capture class-specific structural differences while remaining compatible with neuromorphic SNN classifiers.

## D. SNN Classification Architecture

The spike-encoded inputs are classified using a feedforward spiking neural network (SNN) implemented in PyTorch with the snnTorch library for forward compatibility with GPU acceleration and deployment toolchains, while model verification experiments are cross-checked in Brian2 [26] for neurobiological fidelity [11]. This hybrid development approach ensured that the model design is both biologically interpretable and compatible with modern deep learning infrastructure. The network architecture consisted of:

- Input Layer: 12,800 channels (one per spike-encoded SIFT descriptor).
- Hidden Layers: Two Leaky Integrate-and-Fire (LIF) layers with 512 and 128 neurons, respectively.
- Output Layer: Two LIF neurons corresponding to Pin_OK and Pin_OUT classes.

Each LIF neuron obeyed the standard integrate-and-fire model:

$$\tau_m \frac{du}{dt} = -u(t) + R \cdot I(t) \qquad (2)$$

Where $u(t)$ is the membrane potential, $\tau_m$ the membrane time constant, $R$ the resistance, and $I(t)$ the synaptic input current. A spike is emitted when $u(t)$ exceeded the firing threshold, after which the potential is reset.

The final classification decision is determined by comparing the cumulative spike counts across the two output neurons over the 100-ms simulation window. This decision mechanism not only provided a transparent mapping between neural activity and class output but also enabled low-latency inference, making it suitable for embedded systems.

Sparse spike propagation throughout the network, resulting from the one-spike-per-channel latency encoding, reduces computational load and energy consumption, supporting real-time operation on consumer-grade and embedded hardware.

Figure 7 presents the end-to-end classification pipeline, visually summarising the hybrid SIFT-SNN framework from frame acquisition to binary output classification.

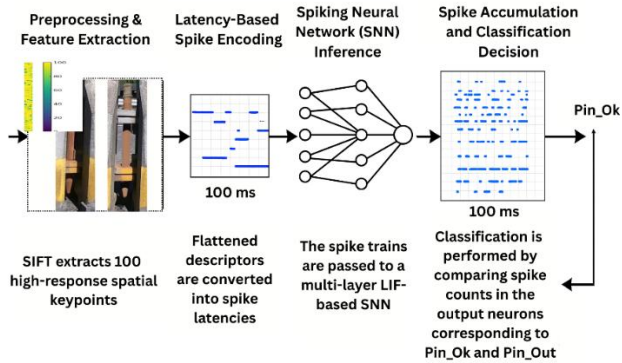

Fig. 7. End-to-end pipeline: video → ROI → SIFT → spike encoding → LIF SNN. The final output is based on spike counts. The system supports low-latency, interpretable, and deployable inspection.

## IV. EXPERIMENTAL SETUP

### A. Dataset Composition, Splits, and Hardware Configuration

The dataset comprised approximately 6,000 annotated frames collected from 20 minutes of high-resolution video recorded during live Movable Concrete Barrier (MCB) transfers on the Auckland Harbour Bridge (AHB). Data acquisition followed the multisensor setup described in Section III-A. A total of 4,500 frames labelled as structurally safe (Pin_OK) based on the correct seating of the safety pins.

Due to operational safety policies, deliberately displacing pins for recording is prohibited. Although occasional unsafe configurations are observed in transit, controlled filming is not permitted. Consequently, approximately 120 base unsafe (Pin_OUT) frames are manually constructed using digital editing and expanded via synthetic augmentation, including spatial distortion, gamma adjustment, occlusion overlays, rotation, and perspective warping, yielding a final set of 1,500 diverse unsafe frames. This methodology provides a practical framework for developing safety models in other infrastructure contexts where the minority class is either rare or unable to be recorded.

All frames processed through a consistent pipeline: grayscale conversion, histogram equalisation, ROI cropping, and pixel normalisation. From each ROI, the top $N = 100$ SIFT descriptors extracted, yielding fixed-length $100 \times 128 = 12{,}800$-dimensional feature vectors. These are converted into latency-coded spike trains for classification by the spiking neural network (SNN).

To preserve the natural class distribution, stratified sampling used to split the dataset:

- Training set: 4,200 frames (70%)
- Validation set: 900 frames (15%)
- Test set: 900 frames (15%).

Figure 8 illustrates both class balance and the separability of SIFT features. The left panel shows the dataset distribution, while the right panel presents a Principal Component Analysis (PCA) projection of descriptors from a representative subset. Although the PCA space shows partial overlap between classes, Pin_OUT samples form distinct peripheral clusters, suggesting that SIFT descriptors capture discriminative structural cues. This supports their suitability for low-latency, spike-based classification.

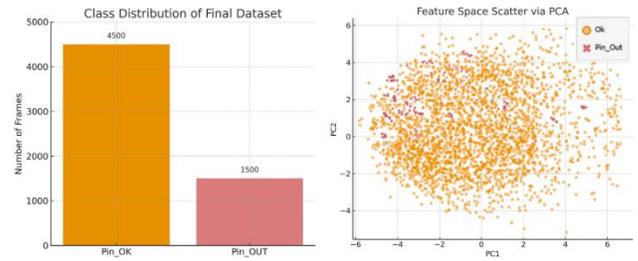

Fig. 8. Dataset visualisation showing (left) class distribution between Pin_OK and Pin_OUT samples, and (right) PCA-based projection of SIFT descriptors from a representative subset. Although some overlap is observed in the transformed feature space, the emergence of peripheral clusters for Pin_OUT highlights the discriminative structure captured by SIFT, justifying its use in spike-based classification.

All experiments are conducted on a personal workstation with the following specifications:

- Processor: Intel Core i7-13650HX (14 cores, 20 threads, up to 4.90 GHz)
- Memory: 64 GB DDR5-5200 MHz RAM
- Storage: 1 TB NVMe SSD
- Graphics: NVIDIA GeForce RTX 4060 (8 GB GDDR6)
- Software Environment: Python 3.10, OpenCV 4.x, PyTorch 2.0, snnTorch

The SNN model is trained using surrogate gradient descent and evaluated using the held-out test set. Metrics included classification accuracy, precision, recall, F1 score, inference latency, and spike activation density.

### B. Training Protocol and Hyperparameters

Each input frame is represented as a 12,800-dimensional spike vector, derived from the top 100 SIFT descriptors (each 128D), latency-encoded over a 100-ms simulation window as described in Section III-C. Higher-magnitude descriptors generated earlier spikes, consistent with the neuromorphic principle that stronger stimuli elicit faster neural responses.

The classification network architecture followed the design in Section III-D, comprising:

- Input layer: 12,800 spike channels
- Hidden layer: 512 leaky integrate-and-fire (LIF) neurons
- Output layer: 2 LIF neurons (Pin_OK, Pin_OUT).

Training is conducted using the Adam optimiser with an initial learning rate of 0.001, decayed by a factor of 0.95 per epoch. The binary cross-entropy (BCE) loss function is used due to the binary classification task. The model reliably converged by epoch 45 (out of 50 total), with no observable overfitting, as confirmed by consistent validation and training loss trajectories.

Overall, the model completed training on the entire training set in approximately 3–4 minutes using an NVIDIA RTX 4060 GPU. This short training time reflects the computational efficiency of the spike-based approach and its suitability for rapid retraining in operational settings.

TABLE 2: TRAINING CONFIGURATION AND HYPERPARAMETERS FOR THE SIFT-SNN MODEL

| Input Feature Size | Time Steps (T) | Batch Size | Epochs | Learning Rate | Optimiser | Loss Function |
|---|---|---|---|---|---|---|
| 12,800 (100 × 128D) | 100 | 64 | 50 (conv. ~45) | 0.001 (decay 0.95/epoch) | Adam | Binary Cross-Entropy |

| Spike Encoding | Hidden Layer | Output Layer | Hardware Used | Training Time |
|---|---|---|---|---|
| Latency coding (magnitude → spike time) | 512 LIF neurons (β = 0.9) | 2 LIF neurons (binary) | Intel i7-13650HX, RTX 4060 GPU, 64 GB RAM | ~3–4 minutes |

### C. Evaluation Metrics and Baseline Models

All models were evaluated using consistent dataset partitions. Reported metrics reflect stable outcomes across repeated trials and confirm the reproducibility of the presented results.

Model performance is evaluated on the 900-frame test set using the following metrics:

- Accuracy: Overall proportion of correctly classified frames.
- Precision (OUT): Proportion of predicted *Pin_OUT* frames that are correct.
- Recall (OUT): Proportion of actual *Pin_OUT* frames correctly identified.
- F1 score: Mean of precision and recall.
- Inference latency: Mean per-frame processing time, including spike encoding and classification.
- Spike activity/activation density: Proportion of active neurons per simulation window, indicating energy efficiency.

For comparisons, Table 3 reports on experimental work including both deep CNNs and non-spiking SIFT-based models.

TABLE 3: COMPARATIVE PERFORMANCE OF CNN BASELINE MODELS AND THE PRESENTED SIFT-SNN MODEL

| Model | Accu-racy (%) | F1 Score (%) | Precision (OUT) (%) | Recall (OUT) (%) | Laten-cy (ms) | Spike Activity | Edge Deploy-able |
|---|---|---|---|---|---|---|---|
| ResNet-50 | 95.1 | 94.0 | 89.0 | 91.0 | 85.0 | Dense | No |
| MobileNetV2 | 91.2 | 90.2 | 86.0 | 88.0 | 42.0 | Moderate | Limited |
| SIFT+MLP | 89.8 | 88.4 | 83.5 | 86.0 | 6.2 | Dense (ANN) | Yes |
| SIFT+SVM | 87.6 | 86.2 | 82.0 | 84.5 | 5.8 | N/A | Yes |
| **SIFT-SNN** | 92.3 | 91.0 | 86.0 | 88.0 | 9.5 | Sparse 8.1% | Yes |

Where:
ResNet-50 … High-capacity CNN optimised for accuracy.
MobileNetV2 … Lightweight CNN optimised for mobile deployment.
SIFT + MLP … Fully connected artificial neural network trained directly on SIFT descriptors.
SIFT + SVM … Classical machine learning classifier for SIFT feature-based baselines.

The proposed SIFT-SNN matched or exceeded MobileNetV2 in all metrics while operating with significantly lower latency and spike activity. While ResNet-50 achieved the highest raw accuracy, its inference latency (85 ms) and dense activation pattern make it unsuitable for real-time embedded deployment.

Unlike CNNs, which require powerful GPUs for sustained real-time performance, the SIFT-SNN can operate on consumer-grade hardware and remains functional in CPU-only configurations with only a minor drop-in frame rate. This makes it well-suited for deployment on barrier transfer machines or similar mobile inspection platforms, where compute resources and energy budgets are constrained.

The inclusion of SIFT + MLP and SIFT + SVM baselines highlights the benefits of the spike-based architecture, as it achieves higher accuracy than traditional descriptor-classifier pipelines while preserving interpretability and enabling compatibility with neuromorphic hardware.

## V. RESULTS

### A. Classification Performance and Latency Efficiency

The proposed SIFT-SNN framework is evaluated on the 900-frame held-out test set derived from the 6,000-frame dataset. Figure 9 presents a composite performance visualisation combining confusion matrix analysis, precision–recall (PR) characteristics, and class-wise metrics.

The confusion matrix (Fig. 9, left) demonstrates strong class discrimination, correctly identifying 641 out of 675 Pin_OK frames and 194 out of 225 Pin_OUT frames. Misclassifications are proportionally balanced across classes, indicating stable boundaries under the 3:1 class imbalance.

The PR curve (Fig. 9, middle) highlights the model's high anomaly detection sensitivity, achieving an average precision of ~0.85 and a recall of ~88% for Pin_OUT. This sensitivity is critical for safety inspections, where false negatives can have operational consequences. The curve profile aligns with neuromorphic systems, in which early spike arrivals often dominate classification.

The class-wise metric chart (Fig. 9, right panel) confirms category stability:

- Pin_OK achieved ~95% across precision, recall, and F1 score.
- Pin_OUT maintained ~86–88% across the same metrics, demonstrating reliable rare-class detection without overfitting to the majority class.

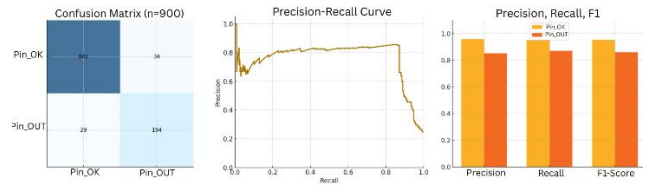

Fig. 9. Composite evaluation of SIFT-SNN classification. Left: confusion matrix; middle: PR curve showing average precision ~0.85; right: class-wise precision, recall, and F1 score. The model maintains high sensitivity under imbalanced data conditions.

To assess real-time suitability, (Fig. 10) benchmarks per-frame inference latency against CNN baselines. ResNet-50 reached 95.1% accuracy but required ~85 ms/frame, making it unsuitable for real-time use; MobileNetV2 is faster (~42 ms) but weaker in anomaly recall.

In contrast, the SIFT-SNN achieved ~9.5 ms/frame on GPU and ~26 ms/frame on CPU, with a spike density of ~8.1%, demonstrating:

- Real-time feasibility on both GPU-equipped and CPU-only systems.
- Energy efficiency from sparse, event-driven computation.
- Operational robustness in detecting anomalies despite constrained hardware budgets.

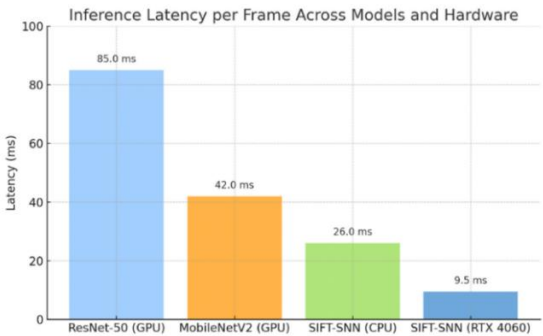

Fig. 10. SIFT-SNN achieves sub-10 ms latency on GPU and under 30 ms on CPU, outperforming CNN baselines in speed and computational efficiency while maintaining competitive accuracy.

### B. Comparative Analysis

While ResNet-50 achieved the highest absolute accuracy (95.1%) and F1 score (94.0%), it incurred a substantial inference cost of ~85 ms per frame (Table 3). Although this latency falls within the range often reported in academic image classification benchmarks (100–200 ms), it is unsuitable for continuous real-time inspection, particularly under CPU-bound or embedded operational constraints.

This limitation was confirmed in the 2024 deep learning demonstration trial for the AHB domain experts [4], where the ResNet-50-based system introduced perceptible processing delays and failed to meet operational requirements for mobile use on the Barrier Transfer Machine (BTM) platform.

By contrast, the SIFT-SNN achieved:

- 92.3% ± 0.8% accuracy and 91.0% F1 score, competitive with MobileNetV2.
- ~9.5 ms/frame latency on GPU and ~26 ms/frame on CPU, enabling smooth frame-by-frame operation without buffering.
- 8.1% spike activity, drastically reducing computational load and supporting energy-aware deployment on low-power hardware.

While MobileNetV2 offered reduced latency relative to ResNet-50 (~42 ms/frame), its recall for rare anomalies is lower, making it less reliable for safety-critical inspection tasks.

The SIFT-SNN, therefore, represents more than an academic proof-of-concept:

- It balances accuracy and latency, outperforming MobileNetV2 in rare-class recall.
- It is hardware-agnostic, performing consistently on both GPU and CPU-only platforms.
- It is energy-efficient, aligning with the constraints of embedded, battery-powered monitoring systems.

These characteristics make the architecture a viable, scalable choice for real-time anomaly detection in transport infrastructure safety monitoring, particularly in resource-constrained, edge-computing contexts.

## VI. DISCUSSION

The SIFT-SNN approach shifts from heavy CNN pipelines toward interpretable, low-power intelligence tailored for resource-constrained infrastructure monitoring. Its objective is to balance accuracy, efficiency, latency, and transparency within a unified framework.

This is achieved by:

- Using SIFT keypoints to provide interpretable spatial features.
- Encoding them as sparse, time-coded spikes for neuromorphic processing.

This design enables real-time anomaly detection (~9.5 ms GPU, ~26 ms CPU) with low activation density (~8.1%), making it efficient and deployable on embedded hardware. Unlike opaque CNNs, the model remains transparent, supporting audit trails for safety teams.

Key practical benefits include:

- Fast, scalable training (<4 minutes) for rapid adaptation.
- Energy-efficient operation on portable or battery-powered devices.
- Reduced post-inspection delays and lower maintenance costs.
- Improved transparency, aligning with regulatory needs for explainable AI.

## VII. CONCLUSION AND FUTURE WORK

This paper introduces a hybrid SIFT-SNN architecture for real-time anomaly detection in transport infrastructure. The approach integrates interpretable SIFT descriptors with latency-coded spiking neural networks, striking a balance between accuracy, efficiency, and transparency for edge deployment.

On a domain-specific dataset from the Auckland Harbour Bridge, the model achieved:

- 92.3% ± 0.8% accuracy
- 0.91 F1 score
- Sub-10 ms GPU latency and ~26 ms CPU latency

- 8.1% average spike activity

These results confirm the method's suitability for resource-constrained environments, where low latency, energy efficiency, and interpretability are critical. Unlike high-complexity CNNs, the SIFT-SNN delivers comparable accuracy at a fraction of the computational cost, while maintaining transparency for audit and compliance.

Limitations remain: the unsafe class relied partly on synthetic samples, and the current scope is limited to binary pin detection. Future work will extend the model by.

- Expanding to multi-class defect detection.
- Deploying on neuromorphic hardware (e.g., Intel Loihi, NVIDIA Jetson, FPGA).
- Exploring unsupervised adaptation via STDP to reduce dependence on labelled data.
- Incorporating sequence-based inference for evolving fault detection.

Overall, the SIFT-SNN offers a reproducible and adaptable blueprint for next-generation, field-deployable safety inspection tools across movable barriers and related transport infrastructure.

ACKNOWLEDGMENT

The Auckland University of Technology supported this research. The authors acknowledge using OpenCV, PyTorch, snnTorch, and Matplotlib for system development and evaluation. The Auckland System Management Alliance facilitated data collection. Special thanks to Gary Bonser, Martin Olive and other staff members for supporting safety briefings and site access during the data collection.

**Disclosure of Interests.**

The authors have no other competing interests to declare relevant to this article's content.